%% file: main.tex
\documentclass[runningheads]{llncs}

\usepackage{eccv}

\usepackage{eccvabbrv}

\usepackage{graphicx}
\usepackage{booktabs}

\usepackage[accsupp]{axessibility}  
\newcommand\mypara[1]{\vspace{0mm}\noindent\textbf{#1}}
\definecolor{mygreen}{RGB}{69, 191, 85}
\definecolor{myred}{RGB}{238, 75, 43}
\usepackage{wrapfig,lipsum,booktabs}

\usepackage[pagebackref,breaklinks,colorlinks,citecolor=eccvblue]{hyperref}

\usepackage{orcidlink}
\usepackage{subcaption}
\usepackage{graphicx}
\usepackage{subcaption}
\begin{document}

\title{InsMapper: Exploring Inner-instance Information for Vectorized HD Mapping} 

\titlerunning{InsMapper}

\author{Zhenhua Xu \and
Kwan-Yee.K.Wong \and
Hengshuang Zhao}

\authorrunning{Zhenhua Xu. et al.}

\institute{The University of Hong Kong}

\maketitle
\begin{center}
Project webpage: \url{https://tonyxuqaq.github.io/InsMapper/}\
\end{center}

\input{sec/0_abstract}    
\input{sec/1_intro}
\input{sec/2_relatedwork}
\input{sec/3_point_correlation}
\input{sec/4_methodology}
\input{sec/5_experiments}
\input{sec/6_conclusion}
\newpage

%
%
\bibliographystyle{splncs04}
\bibliography{main}
\end{document}

%% file: sec/0_abstract.tex
\begin{abstract}
Vectorized high-definition (HD) maps contain detailed information about surrounding road elements, which are crucial for various downstream tasks in modern autonomous vehicles, such as motion planning and vehicle control. Recent works attempt to directly detect the vectorized HD map as a point set prediction task, achieving notable detection performance improvements. However, these methods usually overlook and fail to analyze the important inner-instance correlations between predicted points, impeding further advancements. To address this issue, we investigate the utilization of inner-instance information for vectorized high-definition mapping through transformers, and propose a powerful system named \textbf{InsMapper}, which effectively harnesses inner-instance information with three exquisite designs, including hybrid query generation, inner-instance query fusion, and inner-instance feature aggregation. The first two modules can better initialize queries for line detection, while the last one refines predicted line instances.  InsMapper is highly adaptable and can be seamlessly modified to align with the most recent HD map detection frameworks. Extensive experimental evaluations are conducted on the challenging NuScenes and Argoverse 2 datasets, where InsMapper surpasses the previous state-of-the-art method, demonstrating its effectiveness and generality. 
\keywords{Autonomous Driving \and High-definition Map}
\end{abstract}

%% file: sec/1_intro.tex
\section{Introduction}
\label{sec:intro}
Vectorized high-definition maps (HD maps) play a critical role in today's autonomous vehicles \cite{liu2022vectormapnet,liang2020learning,da2022path}, as they contain detailed information about the road, including the position of road elements (\eg, road boundaries, lane splits, pedestrian crossings, and lane centerlines), connectivity, and topology of the road. Without the assistance of HD maps for perceiving and understanding road elements, unexpected vehicle behaviors may be encountered, such as incorrect path planning results or even vehicle collisions.

Typically, HD maps are created by offline human annotation, which is labor-intensive, inefficient, and expensive. Although there are works proposing to make such an offline process automatic \cite{xu2022centerlinedet,xie2023mv,xu2023rngdet++}, it is not possible to recreate and update the HD map frequently when the road has been modified, such as when a new road is built or an existing road is removed. To address this issue, several recent studies propose to detect local HD maps using an online method \cite{liao2022maptr,liu2022vectormapnet,li2021hdmapnet,liao2023maptrv2,qiao2023end,ding2023pivotnet}. This paper aims to further investigate online HD map detection techniques based on vehicle-mounted sensors (\ie, cameras and LiDARs).

\input{figures/fig_results}
Early online HD map detection works consider road element mapping as a semantic segmentation task in bird's-eye view (BEV) \cite{philion2020lift,ng2020bev,li2022bevformer}, in which road elements are predicted in raster format (\ie, pixel-level semantic segmentation mask). However, rasterized road maps cannot be effectively utilized by downstream tasks of autonomous vehicles, such as motion planning and vehicle control \cite{liu2022vectormapnet,liang2020learning,liu2021role}. Moreover, it is challenging to distinguish instances from the rasterized map, especially when some road elements overlap with each other or when complicated topology is encountered (\eg, road split, road merge, or road intersections). To alleviate these problems, HDMapNet \cite{li2021hdmapnet} proposes hand-crafted post-processing algorithms to better obtain road element instances for vectorized HD maps. However, HDMapNet still heavily relies on rasterized results, restricting it from handling complicated urban scenes. Recently, some work has resorted to predicting vectorized HD maps directly \cite{can2021structured,liu2022vectormapnet,liao2022maptr,shin2023instagram}. A two-stage hierarchical set prediction method is proposed in VectorMapNet \cite{liu2022vectormapnet}. After predicting key points of road elements in the HD map, VectorMapNet sequentially generates intermediate points. Although VectorMapNet is considered to be the first online vectorized HD map detection work, the sequential operation degrades its efficiency and model performance. MapTR \cite{liao2022maptr} further proposes to use DETR \cite{carion2020end,zhu2020deformable} for vectorized HD map detection as a point set prediction problem, and the output point sets are then grouped into road element instances.
Some recent works further improve the performance of MapTR \cite{liao2023maptrv2,ding2023pivotnet,qiao2023end}. Among them, MapTR-V2 achieves state-of-the-art performance, but it overlooks and fails to utilize the essential inner-instance correlations between points to further boost the performance.

In this work, we focus on investigating the utilization of inner-instance information for vectorized high-definition mapping via transformer, and propose a new framework named \textbf{InsMapper}, which can utilize inner-instance point information for improved online HD map detection via three exquisite designs. First, a hybrid query generation module is introduced to better initialize queries for detection. Then, an inner-instance query fusion module is added before the transformer decoder, which fuses inner-instance object queries to refine the initialized line instances. Finally, a transformer-based inner-instance aggregation module is incorporated to further refine the predicted lines. 
Extensive experimental evaluations are conducted on the popular and challenging NuScenes~\cite{caesar2020nuscenes} and Argoverse 2~\cite{wilson2021argoverse} datasets. InsMapper is based on MapTR \cite{liao2022maptr}, and it can be seamlessly modified based on the latest frameworks with more enhanced performance, such as PivotNet \cite{ding2023pivotnet} and MapTR-V2 \cite{liao2023maptrv2}. No matter what the base framework is, InsMapper harvests state-of-the-art performance, surpassing the previous highest solutions. The intuitive performance comparisons are illustrated in Figure~\ref{results}, and our main contributions can be summarized as follows::

\begin{itemize}
    \item We analyze the limitations of existing HD map detection algorithms and propose a new framework named InsMapper, which can effectively utilize the easily overlooked inner-instance point correlation information for accurate and generalizable online HD map detection.
    \item We incorporate three useful modules to leverage inner-instance information, including hybrid query generation, inner-instance query fusion, and inner-instance feature aggregation. The first two modules better initialize queries for detection and the last one refines detected line instances.
    \item We conduct experiments on the challenging NuScenes and Argoverse 2 datasets. InsMapper outperforms all baselines by a large margin. The achieved state-of-the-art results demonstrate its effectiveness and generality.
\end{itemize}

%% file: figures/fig_results.tex
\begin{wrapfigure}{r}{0.5\textwidth}
  \centering
  \vspace{-20pt}
  \includegraphics[width=\linewidth]{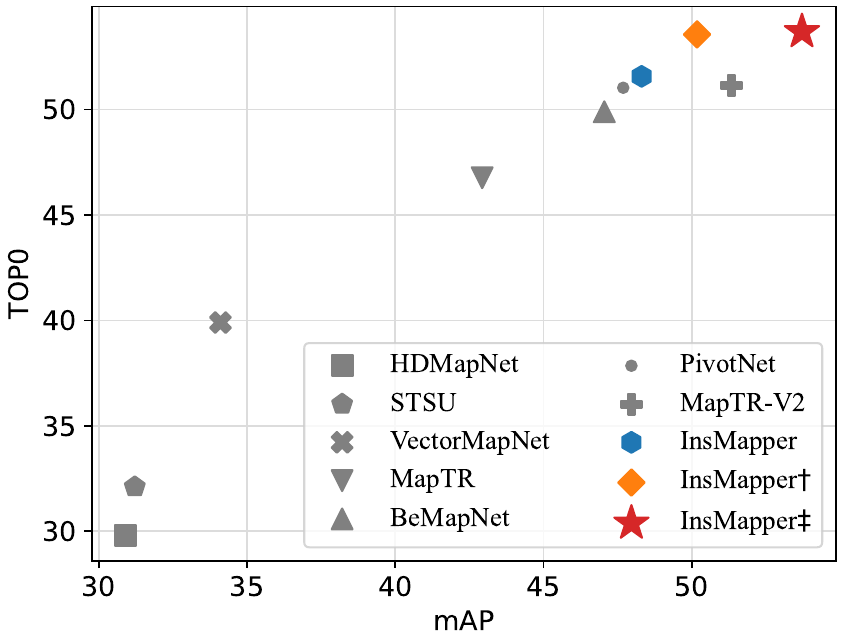}
  \vspace{-20pt}
  \caption{Comparison of vectorized HD map detection methods. All solutions are evaluated on the NuScenes validation set. The x-axis represents the mAP results and the y-axis displays topology level correctness according to \cite{he2018roadrunner}. InsMapper is flexible and adaptable, so it can be seamlessly modified to align with multiple existing frameworks. InsMapper is built on MapTR \cite{liao2022maptr}. $\dagger$ means InsMapper based on more recent framework PivotNet \cite{ding2023pivotnet}, while $\ddagger$ indicates InsMapper based on previous SOTA, MapTR-V2 \cite{liao2023maptrv2}.}
  \label{results}
  \vspace{-20pt}
\end{wrapfigure}

%% file: sec/2_relatedwork.tex
\section{Related Work}
\label{sec:relatedwork}
\mypara{Vector map detection.}
Vector maps use vectors to represent road elements, such as road networks, road curbs, and lane lines. A high-definition map (HD map) is a type of vector map, but a vector map may not qualify as an HD map if it lacks sufficient resolution or fails to provide lane-level information, among other factors. Online HD map detection can benefit from insights gained from a broader scope of vector map detection works,
including road-network detection \cite{bastani2018roadtracer,he2020sat2graph,xu2022rngdet,xu2023rngdet++}, road-curb detection in aerial images \cite{zhxu2021icurb,xu2021topo}, and road lane line detection \cite{homayounfar2018hierarchical,homayounfar2019dagmapper}, etc. DagMapper \cite{homayounfar2019dagmapper} aims to detect vectorized lane lines from pre-built point cloud maps using iterations. However, DagMapper only handles simple topology changes of lane lines and struggles to handle complex road intersections. 
RNGDet++ \cite{xu2023rngdet++} applies DETR along with an imitation learning algorithm \cite{ross2011reduction} for road-network detection, achieving state-of-the-art performance on multiple datasets. Although these methods achieve satisfactory graph detection accuracy, they suffer from poor efficiency due to their iterative algorithms. Given the stringent real-time requirements of autonomous vehicles, they are not suitable for online HD map detection.

\mypara{HD map detection.} 
HD map detection was initially a subtask of BEV detection \cite{philion2020lift,ng2020bev,li2022bevformer,liu2023bevfusion,li2023bi}. Recently, given the importance of HD maps, several works have directly 
 focused on     HD map detection \cite{can2021topology,li2021hdmapnet,mi2021hdmapgen,liu2022vectormapnet,liao2022maptr,xu2022centerlinedet,xie2023mv,shin2023instagram,qiao2023online}. However, most of these works either involve offline HD map creation \cite{xu2022centerlinedet,he2022lane,xie2023mv} or only detect one specific road element \cite{can2021topology,can2021structured,qiao2023online}.  HDMapNet \cite{li2021hdmapnet} is considered to be the first work specifically designed for multiple road element detection. However, HDMapNet outputs rasterized results, requiring complicated hand-crafted post-processing to obtain vectorized HD maps. To address this issue, VectorMapNet \cite{liu2022vectormapnet} is believed to be the first work detecting vectorized HD maps in real-time. However, it consists of two stages, and its efficiency is significantly impacted by sequential operations. In contrast, MapTR \cite{liao2022maptr} uses deformable DETR \cite{zhu2020deformable} to design an end-to-end vectorized HD map detection model, which greatly simplifies the pipeline and delivers better detection performance. Some recent works further improve the performance of MapTR \cite{liao2023maptrv2,ding2023pivotnet,qiao2023end}. Among them, MapTR-V2 achieves state-of-the-art performance on vectorized HD map detection, but it still overlooks and does not investigate the inner-instance correlations between points, limiting further improvements. 

\mypara{Detection by Transformer.}
DETR \cite{carion2020end} is the first end-to-end transformer-based object detection framework. Compared with previous CNN-based methods \cite{girshick2015fast,he2017mask}, DETR removes the need for anchor proposals and non-maximum suppression (NMS), making it simpler and more effective. To address the issue of slow convergence, subsequent works propose accelerating DETR training through deformable attention \cite{zhu2020deformable}, denoising \cite{li2022dn}, and dynamic query generation \cite{wang2022anchor,zhang2022dino,li2023mask}. Intuitively, most of these DETR refinements could be adapted to the HD map detection task since our proposed InsMapper relies on DETR. However, unlike typical object detection tasks in which objects are assumed to be independent and identically distributed (i.i.d.), detected points in HD maps have strong correlations, especially among inner-instance points. This inherent difference makes some refined DETR methods unsuitable to our task and additional investigations are required.

%% file: sec/3_point_correlation.tex
\section{Point Correlation}
\label{sec:pointcorr}
\subsection{Vector Map Decomposition and Sampling}
Let $G$ denote the original vector map label of the scene, which consists of vertices $V$ and edges $E$. The vector map contains multiple classes of road elements, including pedestrian crossings, road dividers, road boundaries, and lane centerlines. Among them, the first three classes of road elements are simple polylines or polygons without intersection points. While lane centerlines have more complicated topology, such as lane split, lane merge, and lane intersection. To unify all vector elements, the vector map is decomposed into simpler shapes (\ie, polylines and polygons) without intersections. Any vertices in the vector map with degrees larger than two (\ie, intersection vertices) are removed from $G$, but incident edges are kept. In this way, a set of simple polylines and polygons without intersections is obtained as $G^*=\{l_i\}_{i=1}^{N^*}$, where $G^*$ is an undirected graph. Each shape $l_i$ is defined as an instance, and $N^*$ denotes the overall number of instances in a vector map.

Following MapTR and MapTR-V2, to enhance the parallelization capacity of the model, each instance is evenly re-sampled with fixed-length points as $l_i=(v_1,...,v_j,...,v_{n_p})$. $l_i$ is ordered from $v_1$ to $v_{n_p}$, with $n_p$ being the number of sampled points for each instance. For polygon instances, $v_1$ is equal to $v_{n_p}$. The pre-processing module is visualized in Figure \ref{pre-processing}.

\input{figures/fig_pre_processing}
\subsection{Inner-instance Correlation}
Unlike conventional object detection tasks, where objects can be approximated as independent and identically distributed (i.i.d.), strong correlations exist between points within the same line instance in the vectorized HD map detection task. The point correlation is visualized in Figure \ref{inner_inter}.

Inner-instance correlation is crucial for point coordinates prediction. Points within the same instance can collaborate by sharing inner-instance information, leading to smoother and more accurate predictions. Without this collaboration, points of the same instance may produce independent predictions, leading to zig-zag or incorrect instances. The inner-instance correlation can serve as additional constraints to facilitate the query initialization and refinement of predicted lines. 
In previous methods, the correlation of points is not correctly analyzed and leveraged, limiting further improvement. In subsequent sections, we introduce InsMapper to more effectively utilize point correlations and improve vectorized map detection performance.

%% file: figures/fig_pre_processing.tex
\begin{figure}[t]
  \begin{minipage}[t]{0.6\linewidth}
    \includegraphics[width=\linewidth]{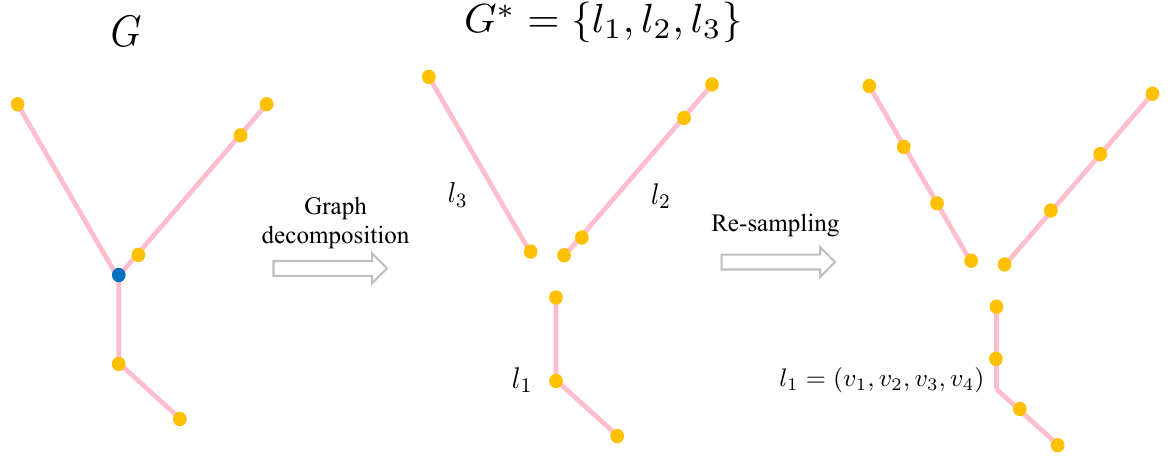}
    \caption{Pre-processing of the vector map. Pink lines represent edges, orange points indicate vertices, and the blue point is the intersection vertex with a degree greater than two. The intersection is removed to simplify the graph, and each obtained instance is then evenly re-sampled into $n_p$ vertices ($n_p=4$ in this example).}
    \label{pre-processing}
    \vspace{-15pt}

  \end{minipage}\rule{1em}{0pt}
  \begin{minipage}[t]{0.35\linewidth}
    \includegraphics[width=\linewidth]{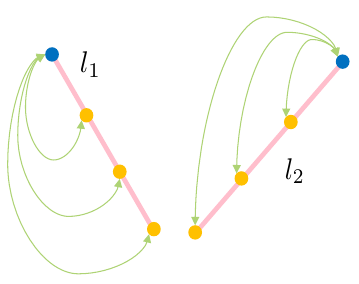}
    \caption{Visualization of inner-instance correlations. Green lines represent the inner-instance correlation between the blue point of an instance and other points within the same instance.}
    \label{inner_inter}
  \end{minipage}
  \vspace{-15pt}

\end{figure}

%% file: sec/4_methodology.tex
\section{Methodology}
\label{sec:methodology}
\input{figures/fig_main_diagram}

\subsection{Overall Framework}\label{1.overall}
Building on MapTR \cite{liao2022maptr}, our proposed InsMapper is an end-to-end trainable framework for online vectorized HD map detection, as illustrated in Figure \ref{diagram}.  \input{figures/fig_query_generation}InsMapper is adaptive and can be seamlessly modified based on the latest frameworks with more enhanced performance, such as PivotNet \cite{ding2023pivotnet} and MapTR-V2 \cite{liao2023maptrv2}. Following BEVformer \cite{li2022bevformer}, InsMapper first projects perspective view camera images into the bird's-eye-view (BEV). After obtaining the BEV features, InsMapper uses deformable attention layers \cite{zhu2020deformable} to process input object queries. Each object query predicts one point, including the object class and the regression point position. To better leverage inner-instance information and further enhance the final detection performance, we introduce three designs in InsMapper: hybrid query generation, which replaces the hierarchical query in MapTR, generating line queries with better diversity; inner-instance query fusion, which fuses generated queries within the same instance based on inner-instance features to further enhance the quality of initialized line queries; and inner-instance feature aggregation, which adds a new inner-instance self-attention module to deformable decoder layers to further aggregate inner-instance features for line refinement.

Different from past works, which treat queries as independently distributed, InsMapper utilizes inner-instance information to fuse object queries and aggregate intermediate features in deformable transformer decoder layers. Otherwise, adjacent vertices may produce independent position regression results, leading to zigzag or even incorrect instances. With the assistance of inner-instance information, the model can better initialize queries for detection and refine the positions of detected points within an instance, significantly enhancing the final performance, as demonstrated in the experiments section.

\subsection{Query Generation}\label{2.query_generation}

In contrast to traditional object detection tasks, where queries are independent and identically distributed (i.i.d.), the points to be detected in our task exhibit significant correlations. Various methods can be used to generate object queries as input for the transformer decoder, such as the basic scheme, hierarchical scheme, and hybrid scheme. All query generation schemes are illustrated in Figure \ref{query_generation}. Initialized queries are organized into line instances. The quality and diversity of the initialized lines are both crucial.

\input{figures/fig_query_initialization}

\mypara{Basic query generation.}
This method is leveraged in the original DETR \cite{carion2020end} for query generation. It assumes that points are i.i.d. and generates queries randomly, without utilizing inner-instance information. Let $N_I$ denote the number of predicted instances, which is obviously larger than $N^*$. Since each instance contains $n_p$ points, the basic object queries consist of $N_I\cdot n_p$ randomly generated i.i.d. queries, resulting in irregular line shapes. Due to the lack of constraints on inner-instance points, the initialized line instances are noisy and have poor quality, which leads to degraded performance. MapTR with the basic query scheme is visualized in Fig. \ref{query_init:1}.

\mypara{Hierarchical query generation.}
To address the aforementioned issue of the basic scheme, a hierarchical query generation scheme is proposed in MapTR \cite{liao2022maptr} and MapTR-V2 \cite{liao2023maptrv2}. Let $Q_{ins}=\{q^I_i\}_{i=1}^{N_I}$ denote instance queries and $Q_{pts}=\{q^P_j\}_{j=1}^{n_p}$ denote point queries. The object queries for HD map detection are then the pairwise addition of $Q_{ins}$ and $Q_{pts}$:
\begin{equation}
    Q=\{q_{i,j}=q^I_i+q^P_j| q^I_i\in Q_{ins}, q^P_j \in Q_{pts}\},
\end{equation}
where $|Q|=|Q_{ins}|\cdot|Q_{pts}|=N_I\cdot n_p$. For each line instance $i$, the instance query $q^I_i$ denotes its uniqueness. By using $q^I_i$ as a bridge, object queries within the same instance can better collaborate with each other for point prediction. But all line instances share the same point queries so all initialized lines tend to have quite similar shapes, degrading initialization diversity, which harms the final detection performance. MapTR with the hierarchical query scheme is visualized in Fig. \ref{query_init:2}.

\mypara{Hybrid query generation.}
The basic scheme does not take into account any information exchange between queries, whereas the hierarchical one generates repetitive line shapes. To address these issues, this paper presents a hybrid query generation method that mitigates the drawbacks of the aforementioned schemes while maintaining appropriate inner-instance information exchange.

Let the instance queries be $Q_{ins} = \{q^I_i\}_{i=1}^{N_I}$, and the point queries be $Q_{pts} = \{q^P_j\}_{j=1}^{N_P} = \{q^P_j\}_{j=1}^{N_I\cdot n_p}$. The point queries are divided into $N_I$ instance groups, and a point query $q^P_j$ is assigned to the $\lceil \frac{j}{n_p} \rceil$-th instance. Consequently, the final object query is the sum of a point query $p^P_j$ and its assigned instance query $p^I_k$, where $k = \lceil \frac{j}{n_p} \rceil$. The object query set can be expressed as:

\begin{equation}
Q = \{q^I_k + q^P_j | q^P_j \in Q_{pts}\}
\end{equation}
where $|Q| = |Q_{pts}| = N_P = N_I\cdot n_p$. In contrast to the hierarchical scheme, each point query in the hybrid scheme is utilized only once to generate the object query, thereby preventing repetitive line shapes. Simultaneously, point queries belonging to the same instance are summed with a shared instance query, establishing the inner-instance connection. The hybrid query generation method can be considered as a combination of the basic and hierarchical schemes, effectively mitigating their respective drawbacks. InsMapper with the hybrid query generation scheme is visualized in Fig. \ref{query_init:3}.

\subsection{Inner-instance Query Fusion}\label{3.Query_fusion}

The input object queries for the transformer decoder are generated from instance queries and point queries. Although the generated queries can leverage some inner-instance information, this information exchange is indirect and inflexible. The instance query is distributed equally among all inner-instance points, while a more precise point-to-point information exchange cannot be realized. As a result, a query fusion module is introduced to further utilize inner-instance information and refine initialized lines.

Let $Q_i = \{q_{i,j}\}_{j=1}^{n_p}$ represent the set of object queries belonging to the $i$-th instance, and $q_{i,j}$ denote the $j$-th point of the $i$-th instance. $q_{i,j}$ is correlated with all other queries in $Q_i$. To better fuse inner-instance object queries, each query is updated by a weighted sum of all queries in $Q_i$ as:
\begin{equation}
    q_{i,j} = f(q_{i,j}, Q_i) = \sum_{k=1}^{n_p} w_{i,j,k} \phi(q_{i,k})
\end{equation}\input{figures/fig_query_generation_init}where $w_{i,j,k}$ demonstrates weights for query fusion. $f(\cdot)$ is the fusion function and $\phi(\cdot)$ is the kernel in case nonlinear transformation is needed. $f(\cdot)$ could be realized by handcraft weights, fully connected layers, or self-attention. In this work, self-attention is applied as $f(\cdot)$.

In conventional object detection tasks, object queries are assumed to be independent and identically distributed (i.i.d.), making query fusion unnecessary. However, in the task of HD map detection, the query fusion module effectively aligns the update of queries and enables each point to pay more attention to neighboring points. It can greatly refine the initialized lines for better query initialization, leading to enhanced performance. Without this module, queries within the same instance cannot be aware of each other, rendering them ``blind''. Poor line initialization tends to have degraded detection scores. The effect of the query fusion module is visualized in Fig. \ref{query_generation_init}.

\subsection{Inner-instance Feature Aggregation}\label{4.feature_agg}

In addition to object query manipulation, InsMapper performs inner-instance feature aggregation for predicted line refinement in the transformer decoder layers by incorporating an additional inner-instance self-attention module. Inspired by \cite{he2022masked}, inner-instance attentions are randomly blocked with probability $\epsilon$ for robustness. The decoder layer of InsMapper is depicted in Figure \ref{ins_self_attn}.

The inner-instance self-attention module resembles the original self-attention module but features a specially designed attention mask. As illustrated in Figure \ref{ins_self_attn}, the attention mask of inner-instance points is set to zero (colored grids), indicating that attention between points within the same instance is allowed. Conversely, for points belonging to different instances (\ie, inter-instance points), the corresponding attention mask values are set to one, blocking attention between them. This method encourages the model to focus more on inner-instance information, resulting in more consistent predictions. To further enhance the robustness of this module, the inner-instance attention (colored grids) has an $\epsilon$ probability of being blocked.

An alternative method involves placing the proposed inner-instance attention module before the cross-attention module. However, the self-attention module before cross-attention should not block inter-instance information. Otherwise, some instances may produce duplicate predictions since they cannot “clearly see” other instances, which is analyzed in the original DETR \cite{carion2020end}. Consequently, implementing inner-instance self attention before the cross-attention module leads to degraded final performance. Thus the proposed inner-instance feature aggregation should be placed after the cross-attention layer. More experiments about the decoder structure are provided in the appendix.

%% file: figures/fig_main_diagram.tex
\begin{figure*}[t]
  \centering
  \includegraphics[width=\linewidth]{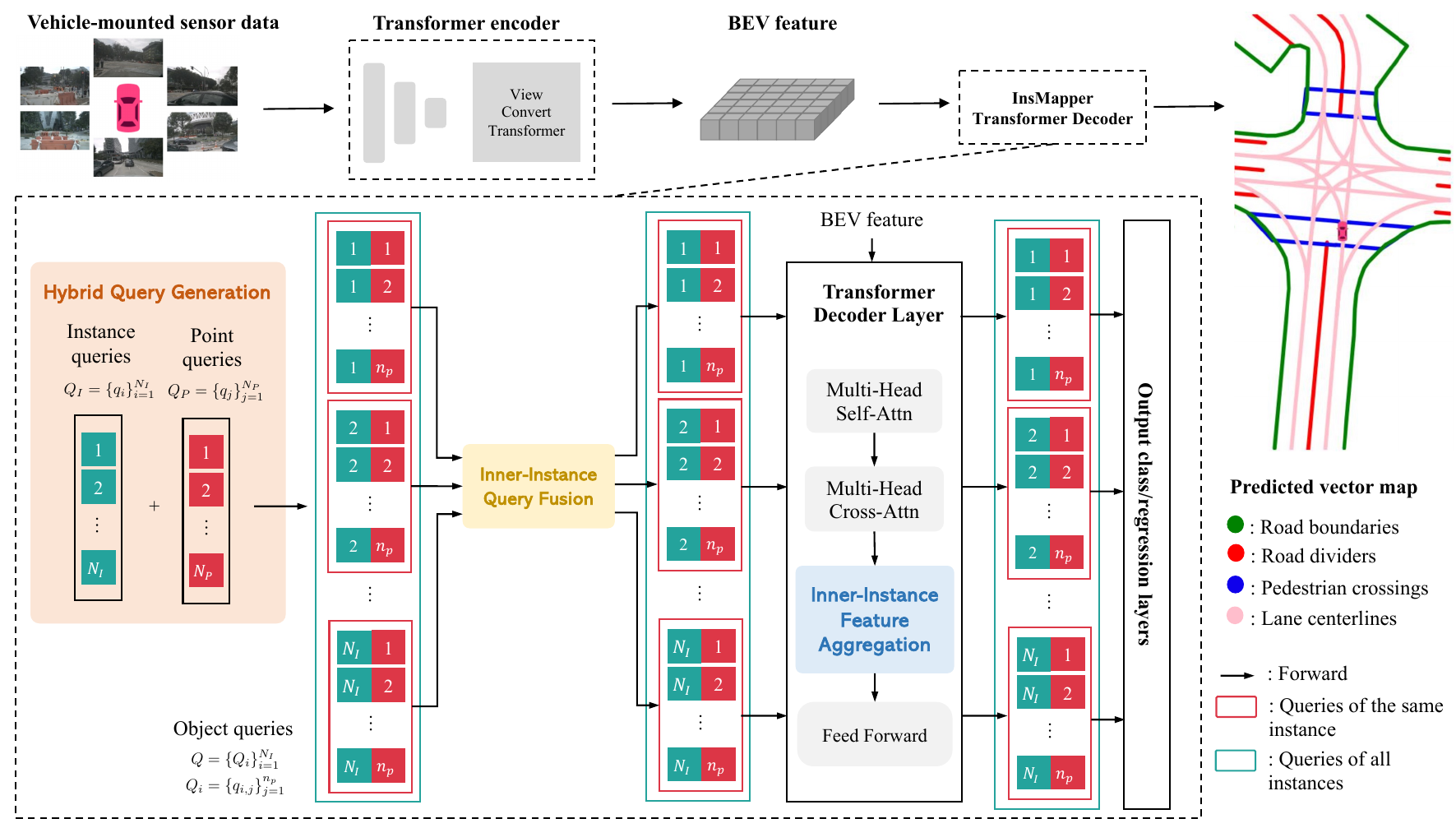}
  \caption{Overall framework. InsMapper is an end-to-end transformer model with an encoder-decoder structure. The transformer encoder projects perspective-view camera images into a bird's-eye view (BEV). Subsequently, the transformer decoder detects vector instances by predicting point sets. To enhance the utilization of inner-instance information, we introduce the following three components: a hybrid query generation scheme (orange module), an inner-instance query fusion module (yellow module), and an inner-instance feature aggregation module (blue module).  The first two modules better initialize queries
for detection and the final one refines detected line instances.}
  \label{diagram}
  \vspace{-10pt}
\end{figure*}

%% file: figures/fig_query_generation.tex
\begin{wrapfigure}{r}{0.6\textwidth}
  \centering
  \includegraphics[width=\linewidth]{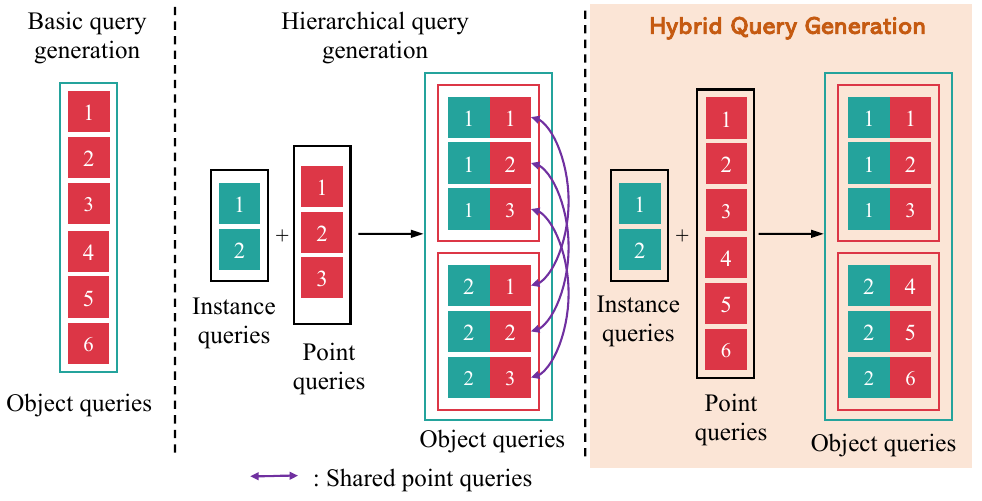}
  \vspace{-20pt}
  \caption{Query generation schemes. For concise visualization, the number of instances $N_I$ is 2, and the number of points per instance $n_p$ is 3. The hybrid scheme can initialize lines with both good diversity and quality.}
  \label{query_generation}
  \vspace{-20pt}
\end{wrapfigure}

%% file: figures/fig_query_initialization.tex
\begin{wrapfigure}{t}{0.6\linewidth}
  \centering
  \vspace{-20pt}
  \begin{subfigure}{0.19\textwidth}
\includegraphics[width=\textwidth]{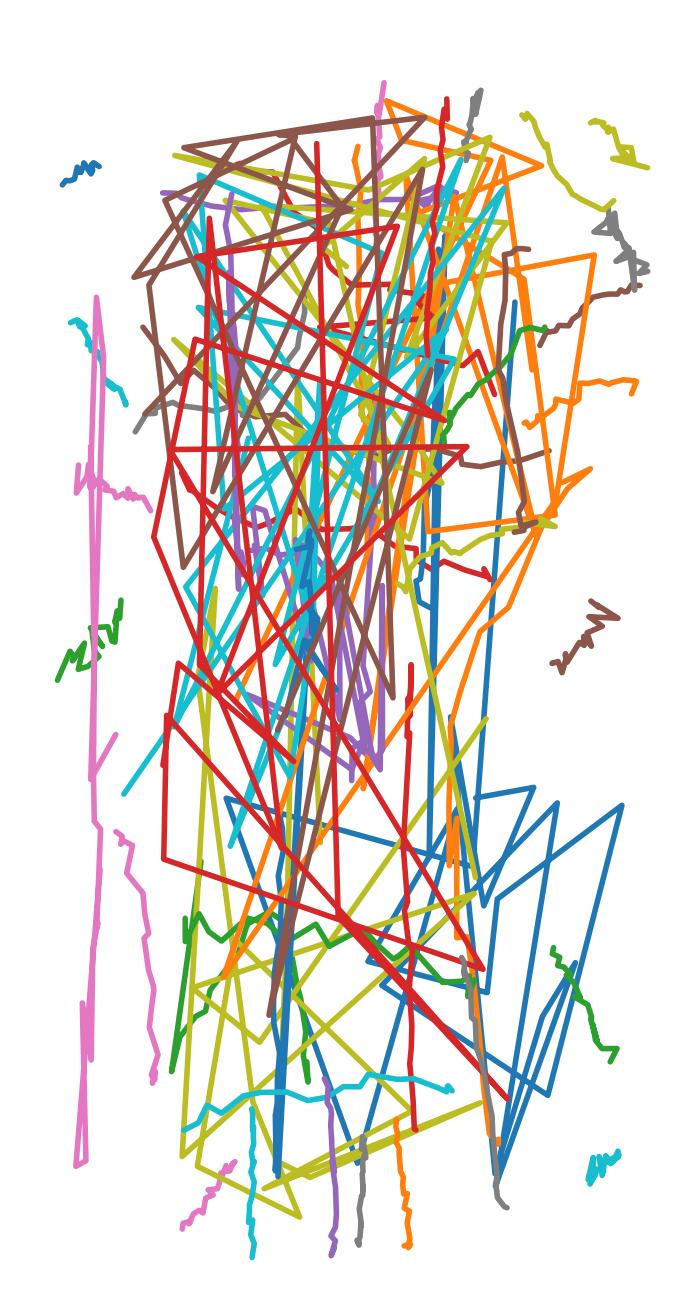}
        \caption{Basic}
        \label{query_init:1}
    \end{subfigure}
    \begin{subfigure}{0.19\textwidth}
    \includegraphics[width=\textwidth]{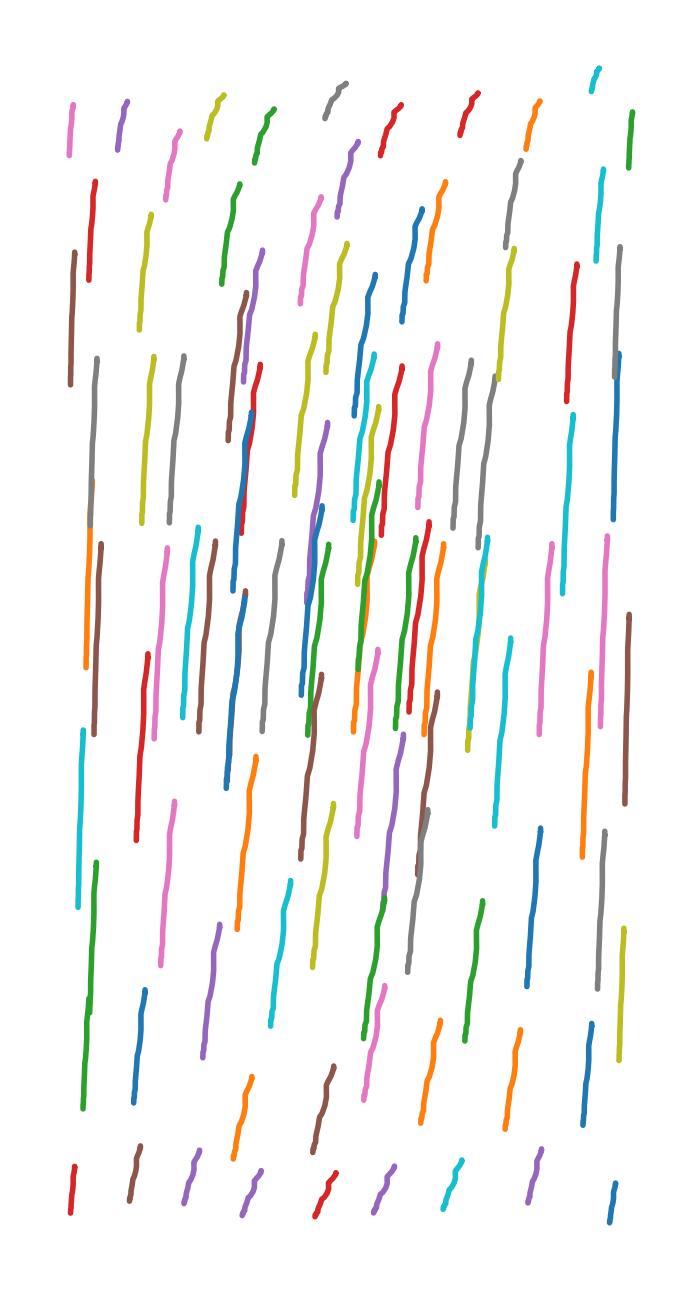}
    \caption{Hierarchical}
        \label{query_init:2}
    \end{subfigure}
    \begin{subfigure}{0.19\textwidth}
    \includegraphics[width=\textwidth]{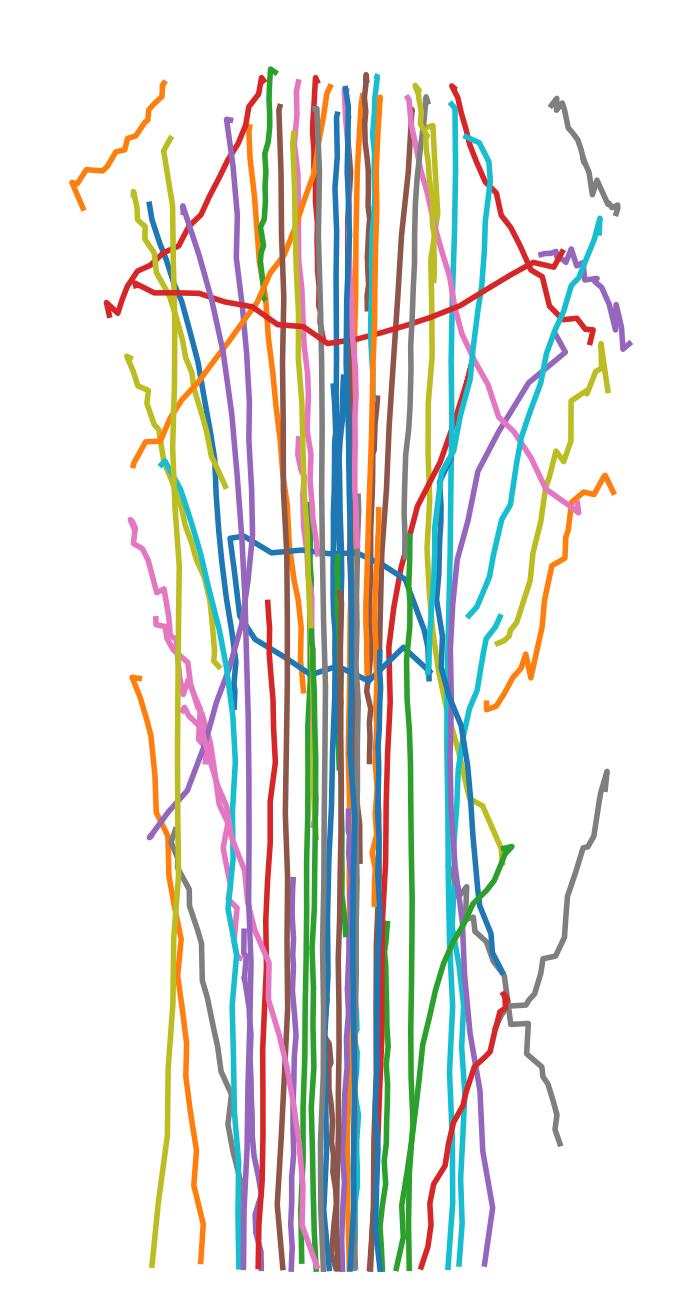}
        \caption{Hybrid}
        \label{query_init:3}
    \end{subfigure}
      \caption{Visualization of initialized lines with different query generation schemes. The quality and diversity of initialized lines are crucial for the final performance.  Each color represents one initialized line. (a) Basic scheme. The basic query scheme generates i.i.d queries, resulting in quite noisy shapes for the initialized lines. (b) Hierarchical scheme. This scheme outputs initialized lines with improved quality but significantly reduced diversity. (c) Hybrid scheme. It generates initialized lines with high quality and good diversity due to the utilization of inner-instance information for query initialization.}
      \label{query_init}
  \vspace{-20pt}
\end{wrapfigure}

%% file: figures/fig_query_generation_init.tex
    
    

\begin{figure}[t]
  \begin{minipage}[t]{0.43\linewidth}
    \begin{subfigure}{0.49\textwidth}
        \includegraphics[width=\textwidth]{imgs/2_baseline.jpg}
        \caption{No fusion}
        \label{query_generation_init:1}
    \end{subfigure}
    \begin{subfigure}{0.49\textwidth}
    
    \includegraphics[width=\textwidth]{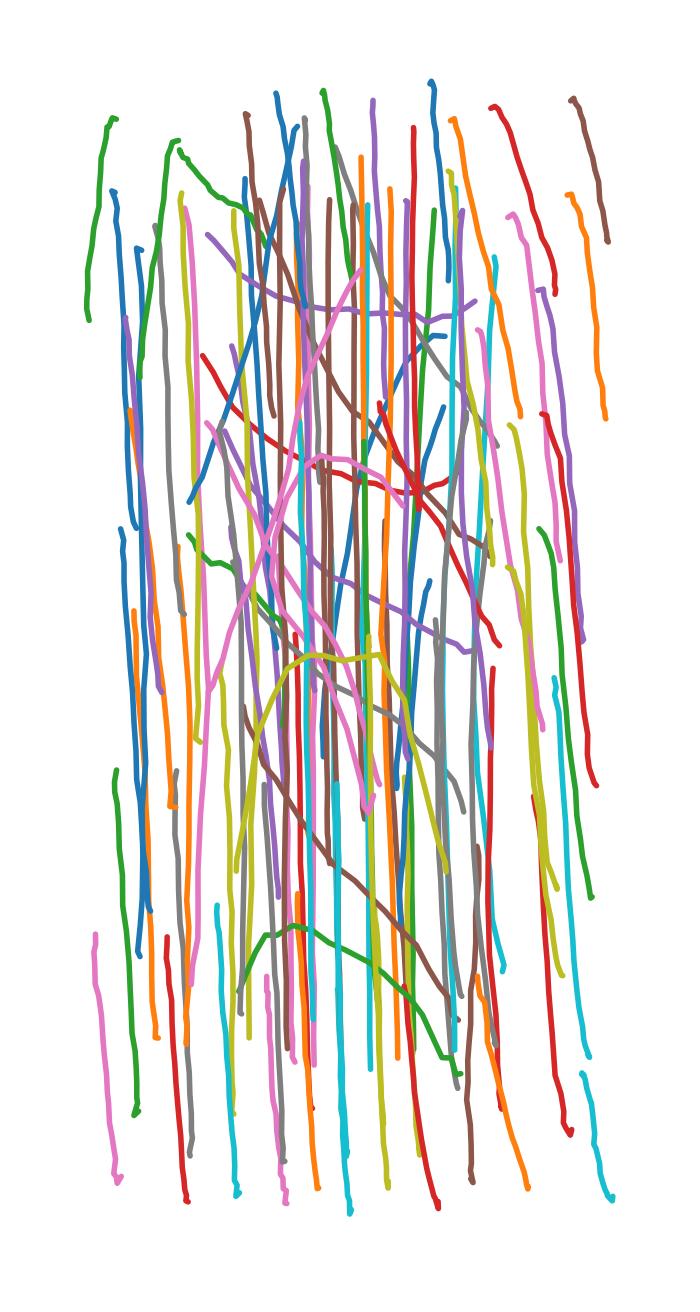}
    \caption{With fusion}
        \label{query_generation_init:2}
    \end{subfigure}
    
      \caption{Visualization of the effect of the inner-instance query fusion module. (a) No query fusion. (b) With query fusion. The query fusion module can better refine the initialized lines based on inner-instance information. In this example, the proposed query fusion module significantly enhances the diversity of initialized line instances.}
\label{query_generation_init}
  \end{minipage}\rule{1em}{0pt}
  \begin{minipage}[t]{0.55\linewidth}
    \includegraphics[width=\linewidth]{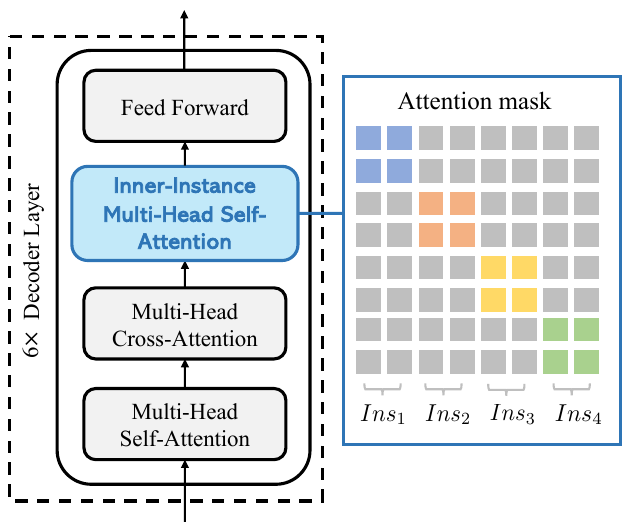}
  \caption{Decoder of InsMapper. An inner-instance multi-head self-attention module is incorporated into decoder layers. In this module, the attention between points belonging to different instances is blocked (grey grids). Only inner-instance attention is allowed (colored grids). Colored grids are randomly blocked with $\epsilon$ probability (set to grey) for robustness. The proposed module can further utilize inner-instance information and refine the predicted lines.}
  \label{ins_self_attn}
  \end{minipage}
  \vspace{-20pt}

\end{figure}

%% file: sec/5_experiments.tex
\section{Experiments}\label{sec:experiments}
\subsection{Experimental Setup}
\mypara{Datasets.}
Extensive experiments are conducted on the popular and challenging NuScenes dataset \cite{caesar2020nuscenes} and the Argoverse 2 dataset \cite{wilson2021argoverse}. These two datasets comprise hundreds of data sequences captured in various autonomous driving scenarios, encompassing diverse weather conditions and time periods. 
The perception range for the X-axis and Y-axis in the BEV is set to $[-15m,15m]$ and $[-30m,30m]$, respectively.
The detected vectorized HD map should encompass four types of road elements simultaneously: road boundary, lane split, pedestrian crossing, and lane centerline. Among these, the lane centerline is a crucial element for ensuring effective vehicle planning and control. Detecting lane centerlines poses greater difficulty compared to the other three elements.

\input{tabs/tab_compare.tex}
\input{tabs/tab_compare_av2}
\input{tabs/tab_compare_no_center}
\input{tabs/tab_compare_directed_graph}
\mypara{Implementation details.}
We perform all the experiments on a machine equipped with 8 RTX-3090 GPUs. During the training phase, all GPUs are utilized, whereas only a single GPU is employed for inference. 
For our proposed InsMapper, we adopt multiple settings akin to the previous SOTA method, MapTR. 
All ablation studies are conducted for 24 epochs, with ResNet50 \cite{he2016deep} as the backbone. For a fair comparison, except for the proposed modules, we keep the experiment settings exactly the same, such as batch size, network depth, the number of input queries, etc.

\input{figures/fig_results_vis}
\mypara{Evaluation metrics.}
Models are assessed using two types of metrics. The first metric, average precision (AP), gauges the instance-level detection performance, employing Chamfer distance for matching predictions with ground truth labels. To ensure a fair comparison, we follow previous works to calculate multiple $\text{AP}_{\tau}$ values with $\tau \in$ \{0.5, 1.0, 1.5\}, and report the average AP. To measure the topology correctness, the metric TOPO that is widely used in past works \cite{he2018roadrunner,he2020sat2graph,he2022lane,xu2023rngdet++} is reported. For all metrics, a larger value indicates better performance. 
\subsection{Performance Comparison}
In this section, we compare InsMapper with previous state-of-the-art methods using the aforementioned evaluation metrics. 
The quantitative comparison results are reported in Table \ref{compare}.
InsMapper is flexible and adaptive, making it seamless to modify it based on existing transformer frameworks. InsMapper is built on MapTR for experiments, we also report the results of InsMapper with different base frameworks, \eg InsMapper$\dagger$ is based on PivotNet \cite{ding2023pivotnet} and InsMapper$\ddagger$ relies on the previous SOTA method MapTR-V2 \cite{liao2023maptrv2}.
Compared to the previous methods, InsMapper improves the AP of all road elements by around 2. The topological metric also sees an improvement of around 1.5. The inference FPS of MapTR and InsMapper are 8.5 and 7.5, respectively.
The evaluation outcomes on the Argoverse 2 dataset \cite{wilson2021argoverse} can be found in Table \ref{compare_av2}. InsMapper outperforms past works with notable enhancements.
Table \ref{compare_no_center} presents the results without centerlines. InsMapper with different bases attains an improvement of more than 2 mAP. In the aforementioned experiments, all road elements are treated as undirected graphs.  Table \ref{compare_nusc_directed} shows the results when centerlines are represented as directed graphs. Based on the results, InsMapper still presents superior results than past works for directed HD map detection. Furthermore, we illustrate the qualitative visualizations in Figure \ref{fig:final_results}. Owing to the effective employment of inner-instance information, InsMapper generates smoother and more precise HD maps compared to previous works.

Consequently, InsMapper demonstrates superior performance compared to prior approaches across multiple datasets, exhibiting outstanding detection outcomes. Concurrently, InsMapper maintains a comparable inference speed to the previous state-of-the-art method. These exceptional experimental results strongly demonstrate the effectiveness and generality of the proposed designs.

\input{tabs/tab_ablation_ways_query_fusion}

\subsection{Ablation Studies}
\mypara{Key component designs.}
We conduct several ablation study experiments to confirm the necessity of the proposed modules. Initially, we incorporate these modules into the previous state-of-the-art baseline model incrementally, and the results are presented in Table \ref{ablation_all}. It is evident that all three modules contribute to performance improvements, thereby validating their necessity. To further investigate the impact of each module's design, we perform comprehensive ablation studies in the following paragraphs.

\mypara{Hybrid query generation.} Compared to basic or hierarchical query generation schemes, the proposed hybrid query generation scheme generates queries with better quality and diversity, due to effectively utilizing inner-instance correlation. The evaluation results of the different query generation schemes are presented in Table \ref{ablation_query_generation}.
These results reveal that the hybrid query generation scheme outperforms its counterparts, thereby demonstrating the soundness of this design.

\mypara{Inner-instance query fusion.} 
Query fusion is proposed to leverage the correlation between inner-instance queries to enhance prediction performance. Various methods can be employed for fusing queries, including mean fusing (\ie, each query is summed with the mean of all queries of the same instance), fusion by the feed-forward network, and fusion by self-attention. The evaluation results are presented in Table \ref{ablation_ways_query_fusion}. The outcomes of the query fusion methods vary significantly, with self-attention-based query fusion achieving the best results. Consequently, self-attention query fusion is incorporated into InsMapper.

\input{tabs/tab_ablation_ins_self_attn}

\mypara{Inner-instance feature aggregation.}
In InsMapper, the inner-instance self-attention module is incorporated into the decoder layers following the cross-attention module. We evaluate InsMapper under four different conditions: without the inner-instance self-attention module (no attention), replace the inner-instance self-attention module with a vanilla self-attention module (vanilla attention), without the random blocking (no random blocking, \ie, $\epsilon$=0), and with the module placed before the cross-attention (change position). The results are shown in Table \ref{ablation_ins_attn}. The outcomes reveal that removing the proposed module, removing the attention mask or altering the position of the proposed module can  impair model performance. This observation confirms the effectiveness of the proposed module design.

%% file: tabs/tab_compare.tex
\begin{table*}[t]
  \caption{Quantitative results of comparison experiments on the NuScenes validation set. Colored numbers show differences between InsMapper and the SOTA baseline under the same experiment setting. ``C'' and ``L'' denote camera and LiDAR, respectively. ``-'' denotes that the result is not available. ``V2-99'' and ``Sec'' correspond to VoVNetV2-99 \cite{lee2020centermask} and SECOND \cite{yan2018second}. $\dagger$ represents InsMapper based on PivotNet \cite{ding2023pivotnet}; $\ddagger$ indicates InsMapper based on MapTR-V2 \cite{liao2023maptrv2}.}
    \vspace{-10pt}
  \renewcommand\tabcolsep{2.9pt} 
  \scriptsize
  \label{compare}
  \centering
  \begin{tabular}{lcccllllll}
    \toprule
    Methods & Epochs & Backbone & Modality & $\text{AP}_{\text{\textit{ped}}}$ & $\text{AP}_{\text{\textit{div}}}$ & $\text{AP}_{\text{\textit{bound}}}$ & $\text{AP}_{\text{\textit{center}}}$ & mAP & TOPO  \\
    \midrule
    HDMapNet  & 30 & Effi-B0 & C &4.41 & 23.73 & 58.17 & 37.29 & 30.90 & 29.79   \\
    STSU     & 110 & R50& C & \hspace{3mm}- & \hspace{3mm}- & \hspace{3mm}- &  31.21 & 31.21  & 32.11 \\
    VectorMapNet     & 130 & R50 & C& 28.66 & 39.74 & 33.06 & 34.93 & 34.10 & 39.90 \\
    MapTR  & 24 & R50 & C& 37.92 & 46.25 & 50.07 & 37.47 & 42.93 & 46.77\\
    MapTR  & 24 & V2-99 & C& 44.57 & 56.25 & 60.58 & 46.30 & 51.92 & 55.16  \\

    MapTR & 24 & R50\&Sec &C\&L &  49.47 & 58.98 & 66.72 & 44.89 & 55.01 & 56.77  \\
    MapTR  & 110 & R50 & C& 49.16 & 59.12 & 58.93 & 47.26 & 53.62 & 59.16 \\
    BeMapNet  & 24 & R50 & C& 43.59 & 54.11 & 51.90 & 38.60 & 47.05&49.90 \\
    PivotNet & 24 & R50 & C& 42.73 & 53.55 & 52.01 & 42.47 & 47.69 & 51.04\\
    MapTR-V2  & 24 & R50 & C& 48.20 & 54.75 & 56.66 & 45.86 & 51.37 & 51.15 \\
    MapTR-V2  & 110 & R50 & C& 58.68 & 65.72 & 67.12 & 56.16 & 61.92 & 63.26 \\

    \midrule
    
    InsMapper & 24 & R50 & C& {44.36} & {53.36} & {52.77} & {42.35} & {48.31} & {51.58}   \\

    InsMapper & 24 & V2-99 & C& {51.16} & {63.71} & {64.47} & {51.40} & {57.68} & {60.00}   \\
    InsMapper & 24 & R50\&Sec& C\&L & {56.00} & 63.42 & {71.61} & 52.85 & {60.97} & 62.51\\
    
    InsMapper & 110 & R50 & C& 55.42 & {63.87} & {65.80} & {54.20} & {59.40} & 66.19 \\
    InsMapper$\dagger$ & 24 & R50 & C& 46.45 & 54.91 & 54.16 & 45.20 & 50.18 & 53.57 \\
    InsMapper$\ddagger$ & 24 & R50 & C& 49.07 & 57.98 & 59.84 & 47.99 & 53.72 & 53.00 \\
    InsMapper$\ddagger$ & 110 & R50 & C& \textbf{62.09} & \textbf{67.60} & \textbf{68.15} & \textbf{58.41} & \textbf{64.06} & \textbf{64.72} \\
    
    \bottomrule
  \end{tabular}
  \vspace{-10pt}

\end{table*}

%% file: tabs/tab_compare_av2.tex
\begin{table*}[!t]
  \caption{Quantitative results of comparison experiments on the Argoverse 2 validation set. $\ddagger$ indicates InsMapper based on MapTR-V2 \cite{liao2023maptrv2}.}
  
    \vspace{-10pt}
  \renewcommand\tabcolsep{3.415pt} 
  \scriptsize
  \label{compare_av2}
  \centering
  \begin{tabular}{lcccllllll}
    \toprule
    Methods & Epochs & Backbone & Modality & $\text{AP}_{\text{\textit{ped}}}$ & $\text{AP}_{\text{\textit{div}}}$ & $\text{AP}_{\text{\textit{bound}}}$ & $\text{AP}_{\text{\textit{center}}}$ & mAP & TOPO  \\
    \midrule
    MapTR  & 6 & R50 & C& 52.88 & 63.68 & 61.18 & 59.73 & 59.37 & 75.79\\
    MapTR-V2  & 6 & R50 & C & 57.16 & 67.96 & 65.25 & 63.20 & 63.39 & 77.94\\
    
    InsMapper & 6 & R50 & C& {{55.61}} & {{66.60}} & {{62.58}} & {{62.67}} & {{61.87}} & {{77.58}}   \\
    InsMapper$\ddagger$ & 6 & R50& C & \textbf{58.91} & \textbf{71.35} & \textbf{66.90} & \textbf{64.70} & \textbf{65.46} & \textbf{79.26}   \\

    \bottomrule
  \end{tabular}
  \vspace{-10pt}

\end{table*}

%% file: tabs/tab_compare_no_center.tex
\begin{table*}[!t]
  \caption{Quantitative results of comparison experiments on the NuScenes validation set without centerline. $\ddagger$ indicates InsMapper based on MapTR-V2 \cite{liao2023maptrv2}.}
    \vspace{-10pt}
  \renewcommand\tabcolsep{7.7pt} 
  \scriptsize
  \label{compare_no_center}
  \centering
  \begin{tabular}{lcccllllll}
    \toprule
    Methods & Epochs & Backbone & Modality & $\text{AP}_{\text{\textit{ped}}}$ & $\text{AP}_{\text{\textit{div}}}$ & $\text{AP}_{\text{\textit{bound}}}$ & mAP  \\
    \midrule
    MapTR &24 & R50 & C & 46.04 & 51.58 & 53.08 &  50.23  \\
    MapTR-V2 &24 & R50 & C & 59.80 &62.40 &62.40 &61.50  \\

    InsMapper &24 & R50 & C & {48.44} & {54.68} & {56.92} & {53.35} \\
    InsMapper$\ddagger$ &24 & R50 & C & \textbf{61.54} & \textbf{65.09} & \textbf{64.62} & \textbf{63.75} \\

    \bottomrule
  \end{tabular}
  \vspace{-10pt}

\end{table*}

%% file: tabs/tab_compare_directed_graph.tex
\begin{table*}[!t]
  \caption{Quantitative results of comparison experiments on the NuScenes validation set.
  Centerlines are directed graphs in this table. $\ddagger$ indicates InsMapper based on MapTR-V2 \cite{liao2023maptrv2}.}
    \vspace{-10pt}
  \renewcommand\tabcolsep{3.455pt} 
  \scriptsize
  \label{compare_nusc_directed}
  \centering
  \begin{tabular}{lcccllllll}
    \toprule
    Methods & Epochs & Backbone & Modality & $\text{AP}_{\text{\textit{ped}}}$ & $\text{AP}_{\text{\textit{div}}}$ & $\text{AP}_{\text{\textit{bound}}}$ & $\text{AP}_{\text{\textit{center}}}$ & mAP & TOPO  \\
    \midrule
    MapTR &24 & R50 & C & 37.39 & 46.61 & 50.69 & 37.22 & 42.98 & 46.98  \\
    MapTR-V2 & 24 & R50 & C & 49.53 & 55.46 &58.48 &53.82 &54.32 & 54.90  \\
    
    InsMapper &24 & R50 & C & {43.24} & {53.77} & {53.89} & {43.22} & {48.53}  & {52.09} \\
    InsMapper$\ddagger$ &24 & R50 & C & \textbf{51.62} & \textbf{57.88} & \textbf{60.90} & \textbf{56.52} & \textbf{56.73}  & \textbf{56.01} \\
    \bottomrule
  \end{tabular}
  \vspace{-20pt}

\end{table*}

%% file: figures/fig_results_vis.tex
\begin{figure*}[t]
  \centering
  \includegraphics[width=\linewidth]{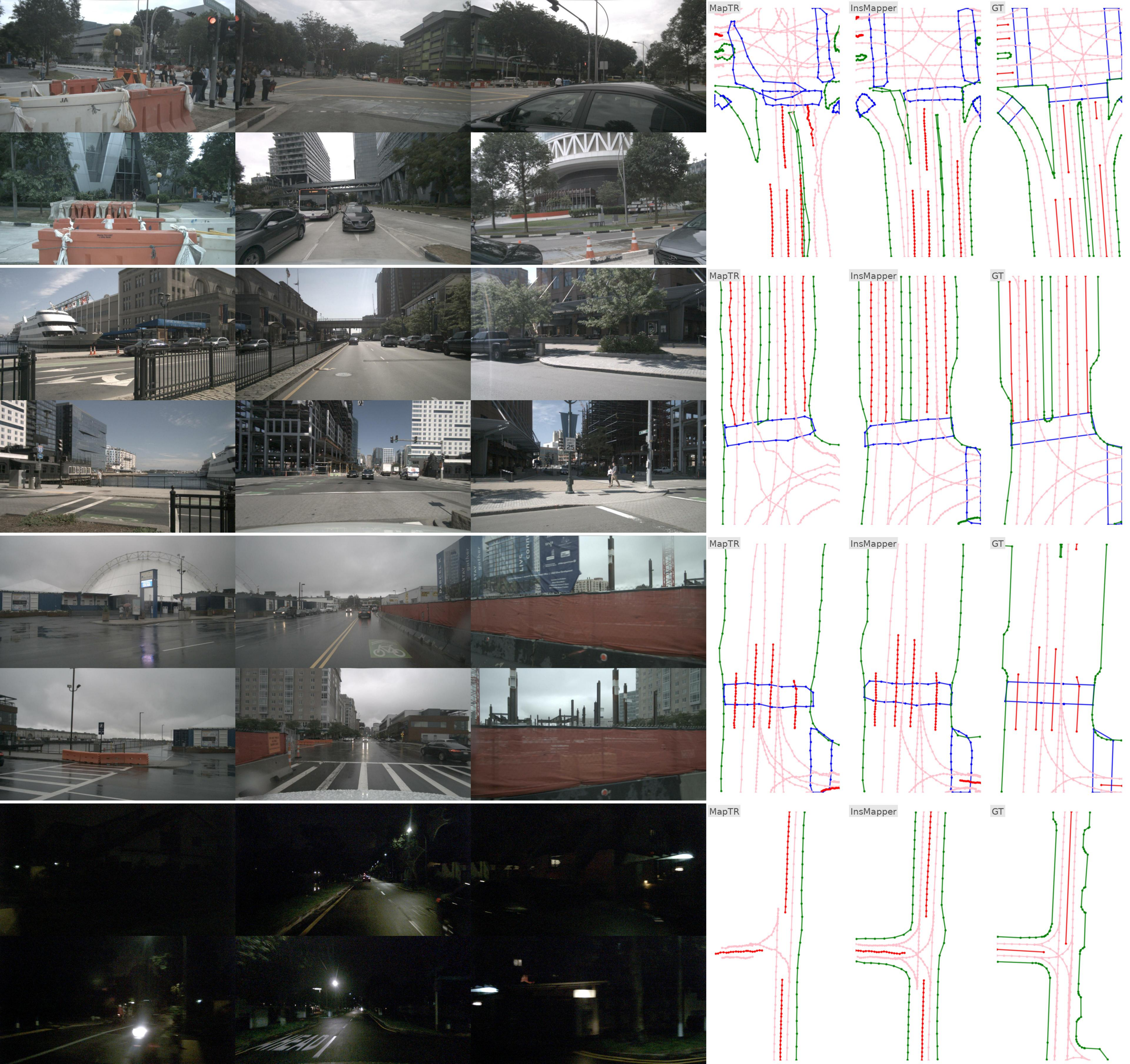}
  \caption{Qualitative visualization. The predicted map contains four classes, i.e., road boundaries (green), lane splits (red), pedestrian crossing (blue), and lane centerlines (pink). InsMapper presents better detection results compared with the past work.}
  \label{fig:final_results}
  \vspace{-20pt}
\end{figure*}

%% file: tabs/tab_ablation_ways_query_fusion.tex
\begin{figure}[t]
\begin{minipage}[t]{.48\linewidth}
    \centering
    \captionof{table}{Ablation studies on key component designs. ``QG'', ``QF'' and ``FA'' represent hybrid query generation, inner-instance query fusion, and inner-instance feature aggregation, respectively. The first row is MapTR while the last row represents InsMapper.}
  \renewcommand\tabcolsep{3pt} 
  \small
  \label{ablation_all}
  \centering
  \begin{tabular}{lllll}
    \toprule
    QG & QF & FA & mAP & TOPO  \\
    \midrule
    && & 42.93 & 46.77  \\
     &\checkmark& & 44.63{\color{mygreen}\scriptsize{($\uparrow$1.70)}}  & 48.44{\color{mygreen}\scriptsize{($\uparrow$1.67)}} \\
     &&\checkmark &  44.31{\color{mygreen}\scriptsize{($\uparrow$1.38)}}  &  48.09{\color{mygreen}\scriptsize{($\uparrow$1.32)}} \\
     
    \checkmark&\checkmark& &  45.42{\color{mygreen}\scriptsize{($\uparrow$2.49)}} & 49.19{\color{mygreen}\scriptsize{($\uparrow$2.42)}}    \\
    
    \checkmark&&\checkmark &  45.45{\color{mygreen}\scriptsize{($\uparrow$2.52)}} & 48.54{\color{mygreen}\scriptsize{($\uparrow$1.77)}}    \\
      
      &\checkmark&\checkmark &  46.93{\color{mygreen}\scriptsize{($\uparrow$4.00)}} & 50.16{\color{mygreen}\scriptsize{($\uparrow$3.39)}}    \\
      
      \checkmark&\checkmark&\checkmark &  48.31{\color{mygreen}\scriptsize{($\uparrow$5.38)}} & 51.58{\color{mygreen}\scriptsize{($\uparrow$4.81)}}    \\
    \bottomrule
  \end{tabular}
  \end{minipage}\hfill
  \begin{minipage}[t]{.48\linewidth}
    \centering
    \captionof{table}{Ablation studies on inner-instance query generation. }
    \renewcommand\tabcolsep{4pt} 
  \small
  \label{ablation_query_generation}
  \centering
  \begin{tabular}{lll}
    \toprule
    
    Query Scheme & mAP & TOPO\\
    \midrule
    
     InsMapper-basic & 46.62  & 49.39\\
     InsMapper-hierarchical &  46.93 &  50.16 \\
     InsMapper-hybrid &  48.31 & 51.58  \\
      \bottomrule
      
  \end{tabular}

    \captionof{table}{Ablation studies on inner-instance query fusion.}
    \renewcommand\tabcolsep{10pt} 
  \small
  \label{ablation_ways_query_fusion}
  \centering
  \begin{tabular}{lll}
    \toprule
    
    Query Fusion & mAP & TOPO\\
    \midrule
    
     No fusion &   45.45 & 48.54  \\
     
     Mean &  43.34 & 46.00 \\
     
    Feed-Forward & 47.80 & 50.93 \\
    
      Self-attention &  \textbf{48.31} &  \textbf{51.58} \\
      \bottomrule
      
  \end{tabular}
  \end{minipage}
  
    \vspace{-10pt}
\end{figure}

%% file: tabs/tab_ablation_ins_self_attn.tex
  \begin{wraptable}{r}{0.5\textwidth}
    \centering
    \vspace{-30pt}
    \captionof{table}{Ablation studies on inner-instance feature aggregation.}
   \renewcommand\tabcolsep{4.8pt} 
  \label{ablation_ins_attn}
  \centering
  \begin{tabular}{lll}
    \toprule
    Method & mAP & TOPO  \\
    \midrule
    
     No attention & 45.42 & 49.19 \\

     Vanilla attention & 46.23 & 49.51\\
     
     No random blocking & 47.67  & 51.49 \\
     
    Change position &   45.24  & 48.16  \\
    
      InsMapper &  \textbf{48.31} & \textbf{51.58}   \\
    
    \bottomrule
  \end{tabular}
  \vspace{-10pt}

\end{wraptable}

%% file: sec/6_conclusion.tex
\section{Conclusion}\label{sec:conclusion}
In this paper, we introduced InsMapper, an end-to-end transformer-based model for on-the-fly vectorized HD map detection. InsMapper surpasses previous works by effectively leveraging inner-instance information to improve detection outcomes. Three exquisite designs have been proposed to utilize inner-instance information, including hybrid query generation, inner-instance query fusion, and inner-instance feature aggregation. The first two modules improve the quality of initialized line instances for detection, while the last module effectively refines the detected lines. The superiority of InsMapper is well exhibited through experiments on the challenging NuScenes and Argoverse 2 datasets. The evaluation results demonstrate the effectiveness and generality of InsMapper, making it a promising solution for the vectorized HD map detection task.